%% file: main.tex
\def\year{2018}\relax
\documentclass[letterpaper]{article} 
\usepackage{aaai18}  
\usepackage{times}  
\usepackage{helvet}  
\usepackage{courier}  
\usepackage{url}  
\usepackage{graphicx}  
\usepackage{pifont}
\usepackage{multirow}
\usepackage{csquotes}
\begin{document}
%
\title{DLPaper2Code: Auto-generation of Code from Deep Learning Research Papers}
\author{Akshay Sethi \thanks{Akshay Sethi interened at IBM Research, India during this work.}\\ IIIT Delhi
	\AND Anush Sankaran, Naveen Panwar, Shreya Khare, Senthil Mani \\ IBM Research, India}
\maketitle

\begin{abstract}
With an abundance of research papers in deep learning, reproducibility or adoption of the existing works becomes a challenge. This is due to the lack of open source implementations provided by the authors. Further, re-implementing research papers in a different library is a daunting task. To address these challenges, we propose a novel extensible approach, DLPaper2Code, to extract and understand deep learning design flow diagrams and tables available in a research paper and convert them to an abstract computational graph. The extracted computational graph is then converted into execution ready source code in both Keras and Caffe, in real-time. An \textit{arXiv}-like website is created where the automatically generated designs is made publicly available for $5,000$ research papers. The generated designs could be rated and edited using an intuitive drag-and-drop UI framework in a crowdsourced manner. To evaluate our approach, we create a simulated dataset with over $216,000$ valid design visualizations using a manually defined grammar. Experiments on the simulated dataset show that the proposed framework provide more than $93\%$ accuracy in flow diagram content extraction.
\end{abstract}

\input{1_introduction}

\input{3_Proposed_Approach}
\input{3_Data_Simulation}

\input{4_Experimental_Study}

\input{7_Conclusion}

\bibliographystyle{aaai}
\bibliography{sigproc} 

\end{document}


\title{DLPaper2Code: Auto-generation of Code from Deep Learning Research Papers}
\author{Supplemnetary Material}
\date{}

\maketitle

\section{Introduction}
This supplementary material provides additional visualizations and results of the main paper.

\section{Evaluation on Simulated Dataset}	

An example of Keras and Caffe visualization is shown in Figure~\ref{fig:kerascaffe}.

\begin{figure}[ht]
	\centering
	\includegraphics[width=3.2in]{caffe-keras.pdf}
	\caption{An example of Keras and Caffe visualization of a DL model design. As it can be manually observed the visualizations are highly varied in node size, color, shape, and the information.}
	\label{fig:kerascaffe}
\end{figure}

\subsection{Extracting Content from a Keras Visualization}
The step-by-step intermediate result on extracting the flow information from a Keras visualization is shown in Figure~\ref{fig:keras_vis}.

\begin{figure*}[ht]
	\centering
	\includegraphics[width=5.5in]{keras_vis.pdf}
	\caption{Illustration of the proposed flow detection algorithm for a Keras visualization.}
	\label{fig:keras_vis}
\end{figure*}

\subsection{Extracting Content from a Caffe Visualization}
The step-by-step intermediate result on extracting the flow information from a Caffe visualization is shown in Figure~\ref{fig:caffe-viz}.

\begin{figure*}[ht]
	\centering
	\includegraphics[width=5.5in]{caffe-viz.pdf}
	\caption{Illustration of the proposed flow detection algorithm for a Caffe visualization.}
	\label{fig:caffe-viz}
\end{figure*}

\subsection{Figure Type Classification Performance}
In this experiment, a binary NNet classifier with two hidden layers of size [$1024$, $256$] is trained on \textit{fc2} features of \textit{VGG19} model to differentiate $216,000$ simulated DL visualizations from a set of $28,120$ other kind of diagrams often available in research papers (scraped from PDF). The whole dataset is split as $60\%$ for training, $20\%$ validation, and $20\%$ for testing, making it a total of $195,296$ images for training and $48,824$ images for testing. The performance of the NNet classifier is compared with six different classifiers. The confusion matrix of the performance of each classifier is shown below, where the rows indicate the ground truth label and the columns indicate the predicted label. Also, the first row is the non-deep learning model and the second row corresponds to deep learning model designs.

\noindent\textbf{Naive Bayes}:

[[ 25844    867]

[  4552 200648]]

\noindent\textbf{Decision Tree}:

[[ 24883   1828]

[  1514 203686]]

\noindent\textbf{Logistic Regression}:

[[ 26565    146]

[     0 205200]]

\noindent\textbf{ Random Decision Forest (RDF)}

[[ 26077    634]

[     9 205191]]

\noindent\textbf{SVM (RBF Kernel)}

[[ 26465    246]

[     2 205198]]

\noindent\textbf{NNet classifier}

[[ 26563    148]

[     7 205193]]

\section{Results on Deep Learning Scholarly Papers}

\subsection{Coarse level Figure Type Classification}

To evaluate the coarse level binary classification, a 2 hidden layer NNet was trained on the \textit{fc2} features obtained from the $30,987$ images extracted from research papers. The whole dataset is split as $60\%$ for training, $20\%$ validation, and $20\%$ for testing and the results are computed for seven different classifiers.  The confusion matrix of the performance of each classifier is shown below, where the rows indicate the ground truth label and the columns indicate the predicted label. Also, the first row is the non-deep learning model and the second row corresponds to deep learning model designs.

\noindent\textbf{Naive Bayes}:

[[16287 10424]

[  135  1363]]

\noindent\textbf{Decision Tree}:

[[19839  6872]

[  363  1135]]

\noindent\textbf{Logistic Regression}:

[[22796  3915]

[  241  1257]]

\noindent\textbf{ Random Decision Forest (RDF)}

[[22056  4655]

[  155  1343]]

\noindent\textbf{SVM (RBF Kernel)}

[[22756  3955]

[  204  1294]]

\noindent\textbf{NNet classifier}

[[23031  3680]

[  197  1301]]

\subsection{Fine level Figure Type Classification}

Further, to evaluate the fine level, five-class, figure type classification, the $2,871$ DL design flow diagrams were manually labelled. The distribution of figures were as follows: (i) Neurons plot: $587$ figures, (ii) 2D box: $1,204$, (iii) Stacked2D box: $408$, (iv) 3D box: $562$, and (v) Pipeline plot: $110$. A $60-20-20$ train, validation, and test split is performed to train the NNet classifier in comparison with the six other classifier to perform this five class classification. The confusion matrix of the performance of each classifier is shown below, where the rows indicate the ground truth label and the columns indicate the predicted label. Also, the first row is 2D box, second row is 3D box, third row is Stacked2D box, fourth row is Neurons plot, and fifth row is Pipeline plot.

\noindent\textbf{Naive Bayes}:

[[83 29 15 29 85]

[19 44  7 13 36]

[28  7 57  8 22]

[ 3  2  1  7  3]

[ 5 10  4 16 41]]

\noindent\textbf{Decision Tree}:

[[132  27  33  10  39]

[ 29  49  14   2  25]

[ 30   8  71   3  10]

[  7   3   0   3   3]

[ 27  14   3   5  27]]

\noindent\textbf{Logistic Regression}:

[[183  22  18   1  17]

[ 12  81   5   2  19]

[ 23   4  92   0   3]

[  8   0   0   4   4]

[ 17  19   4   3  33]]

\noindent\textbf{ Random Decision Forest (RDF)}

[[210  13  15   0   3]

[ 47  67   2   0   3]

[ 36   4  82   0   0]

[ 11   2   0   3   0]

[ 35  21   3   0  17]]

\noindent\textbf{SVM (RBF Kernel)}

[[178   6  19   5  15]

[ 32  69   3   2   8]

[ 19   1  93   1   2]

[  9   3   2  12   1]

[ 21  14   8   3  48]]

\noindent\textbf{NNet classifier}

[[184  27   7   1  22]

[ 10  87   3   0  19]

[ 16   8  93   0   5]

[  6   2   0   4   4]

[  8  22   3   0  43]]

%% file: 1_introduction.tex
\section{Introduction}
The growth of deep learning (DL) in the field of artificial intelligence has been astounding in the last decade with about $35,800$ research papers being published since $2016$\footnote{\url{https://scholar.google.co.in/scholar?as_sdt=1,5\&q=\%22deep+learning\%22\&hl=en\&as_ylo=2016\&as_vis=1}}. Keeping up with the growing literature has been a real struggle for researchers and practitioners. In one of the recent AI conferences, NIPS $2016$, the maximum number of papers submitted ($\sim685/2500$) were in the topic, ``Deep Learning or Neural Networks". However, a majority of these research papers are not accompanied by their corresponding implementations. In NIPS $2016$, {\bf only} $101/ 567$ ($\sim18\%$) papers made their source implementation available\footnote{\url{https://www.kaggle.com/benhamner/nips-papers}}. Implementing research papers takes at least a few days of effort for software engineers assuming that they have limited knowledge in DL~\cite{sankaran}.

Another major challenge is the availability of various libraries in multiple programming langauges to implement DL algorithms such as Tensorflow~\cite{tensorflow2015}, Theano~\cite{theano2012}, Caffe~\cite{jia2014caffe}, Torch~\cite{torch}, MXNet~\cite{chen2015mxnet}, DL4J~\cite{DL4J}, CNTK~\cite{CNTK} and wrappers such as Keras~\cite{chollet2015keras}, Lasagne~\cite{lasagne}, and PyTorch~\cite{pytorch}. The public implementations of the DL research papers are available in various libraries offering very little interoperability or communication among them. Consider a use-case for a researcher working in ``image captioning", where three of the highly referred research papers for the problem of image captioning\footnote{\url{https://competitions.codalab.org/competitions/3221#results}} are:
\begin{enumerate}
	\item Show and Tell~\cite{vinyals2015show}: Original implementation available in Theano; \url{https://github.com/kelvinxu/arctic-captions} 
	\item NeuralTalk2~\cite{karpathy2015deep}: Original implementation available in Torch; \url{https://github.com/karpathy/neuraltalk2}
	\item LRCN~\cite{donahue2015long}: Original implementation available in Caffe; {\url{http://jeffdonahue.com/lrcn/}}
\end{enumerate}

\noindent As the implementations are available in different libraries, a researcher cannot directly combine the models. Also, for a practitioner having remaining of the code-base in Java (DL4J) directly leveraging either of these public implementations would be daunting. Thus, we highlight two highly overlooked challenges in DL:
\begin{enumerate}
	\item Lack of public implementation available for existing research works and thus, the time incurred in reproducing their results
	\item Existing implementations are confined to a single (or few) libraries limiting portability into other popular libraries for DL implementation.
\end{enumerate}
\begin{figure*}[ht]
	\centering
	\includegraphics[width=6.8in]{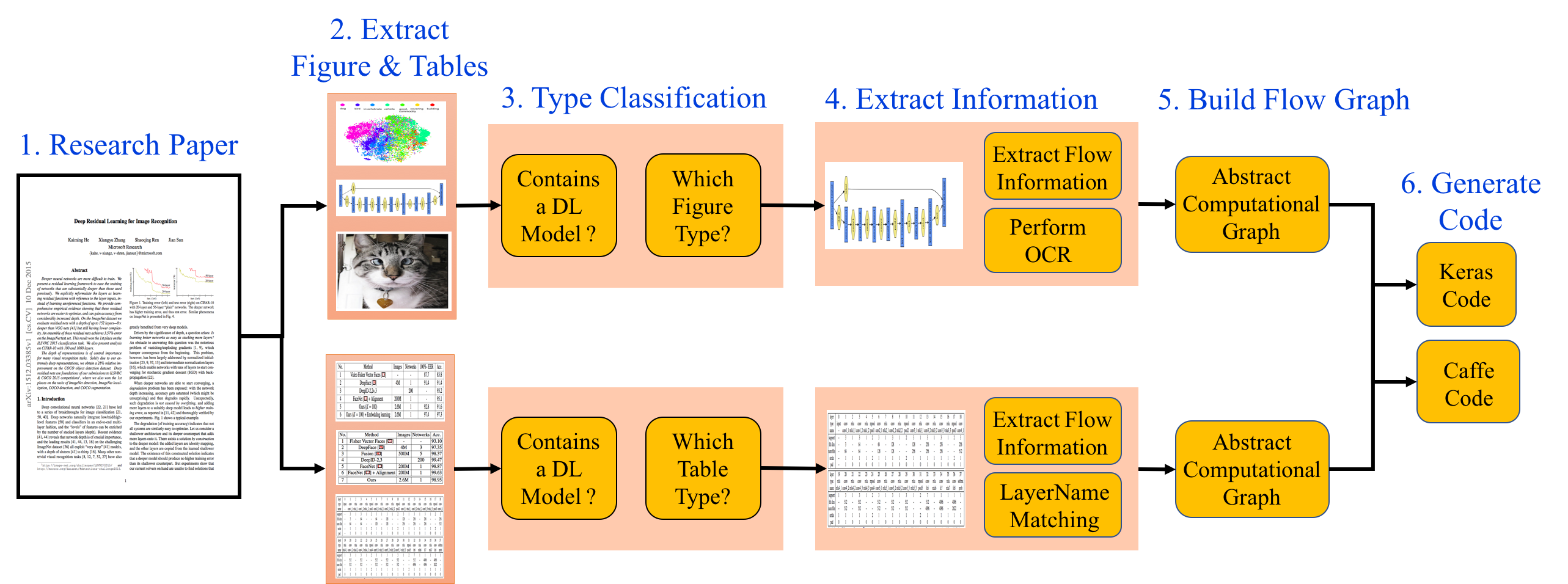}
	\caption{The architecture of the proposed creative system that extracts and understands the flow diagram from a deep learning research paper and generates an execution ready deep learning code in two differcaffent platforms: Keras and Caffe.}
	\label{fig:arch}
\end{figure*}

\noindent We observed that most of the research paper explains the DL model design either through a figure or a table. Thus, in this research we propose a novel algorithm that automatically parses a research paper to extract the described DL model design. The design is represented as an abstract computational graph which is independent of the implemenation library or language. Finally, the source code is generated in multiple libraries from this abstract computational graph of the DL design. The results are shown by automatically generating the source code of $5,000$ \textit{arXiv} papers both in CAFFE (prototxt) and Keras (python). However, evaluating the generated source code is debatable due to the lack of ground truth. To overcome this challenge, we simulated a large image dataset of $216,000$ valid DL model designs in both Caffe and Keras. To generate DL visualizations, we manually defined a grammar for DL models. As these visualizations are highly varying, they are comparable to the figures present in research papers. Thus, the major research contributions are:
\begin{enumerate}
	\item a technique to automatically understand a DL model design by parsing the figures and tables in a research paper,
	\item generate source code both in both Keras and Caffe from the abstract computation graph of a DL design,
	\item automatically generate design for $5,000$ \textit{arXiv} papers and build a UI system for editing them the crowdsourced way,
	\item on a simulated dataset of $216,000$ DL model visualizations using a manually defined grammar, evaluate the proposed approach to achieve more than $95\%$ accuracy.
\end{enumerate}

The rest of the paper is organized as follows: Section 2 explains the entire proposed approach for auto generation of DL source code from research paper, Section 3 talks about the simulated dataset and its experimental performance of the individual components of the proposed approach, Section 4 discusses the experimental results on $5,000$ \textit{arXiv} DL papers, and Section 5 concludes this paper with some discussion on our future efforts.

%% file: 3_Proposed_Approach.tex
\section{Proposed Approach}
Consider a state-of-art paper DL paper~\cite{szegedy2017inception} published in AAAI 2017, which explains the DL design model through figures, as shown in Figure~\ref{fig:flow2}(a). Similarly, in the AAAI 2017 paper by~\cite{parkhi2015deep}, the DL model design was explained using a table.
Thus, given the PDF of a research paper in deep learning, the proposed DLPaper2Code architecture consists of five major steps, as shown in Figure~\ref{fig:arch}: (i) Extract all the figures and tables from a research paper. Handling the figure and table content are done independently, although the follow a similar pipeline, (ii) As there could be other descriptive image and results tables in a paper, classify each figure or table whether it contains a DL model design. Also, perform a fine grained classification on the type of figure or table used to describe the DL model, (iii) Extract the flow and the text information from the figures and tables, independently, (iv) Construct an abstract computational graph which is independent of the implementation library, and (v) generate source code in both Caffe and Keras from the computational graph.


\begin{figure*}[ht]
	\centering
	\includegraphics[width=7in]{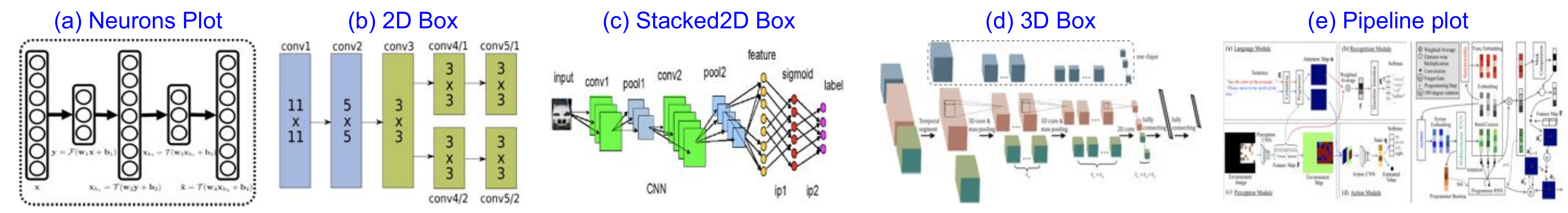}
	\caption{Characterizing the DL model designs available in research papers and grouping them into five different categories.}
	\label{fig:types}
\end{figure*}

\subsection{Characterizing Research Papers}
We observed that in a research paper the DL design is mostly explained using figures or tables. Thus, it is our assertion that by parsing the figure, as an image, and the table content, the respective novel DL design could be obtained. The primary challenges with the figures in research papers is that the DL design figures typically do not follow any definition and show extreme variations. Similarly, tables can have different structures and can entail different kind of information.

We manually observed more than $30,000$ images from research papers and characterized the DL model deisgn images into five broad categories, as shown in Figure~\ref{fig:types}. The five types are: (i) Neurons plot: the classical representation of a neural network with each layer having circular nodes inside them, (ii) 2D Box: each hidden layer is represented as a 2D rectangular box, (iii) Stacked2D Box: each layer is represented as a stack of 2D rectangular boxes, describing the depth of the layer, (iv) 3D Box: each hidden layer is represented as a 3D cuboid structure, and (v) Pipeline plot: along with the DL model design, the entire pipeline and mostly some intermediate results of image/ text is shown as well. Similarly, based on the representation, tables can be classified as, (i) row-major table: where the DL model design flows along the row~\cite{springenberg2014striving}, and (ii) where the DL model design flows along the column~\cite{parkhi2015deep}. It is essential to account for these variations in the proposed pipeline, as they indicate the DL design flow represented in the paper. Following this assumption, the proposed approach does not identify a DL design flow that is neither in a table nor in a figure.

\subsection{Extracting Figures and Tables}
Extracting visual figures from a PDF document, especially a scholarly report is a well studied problem~\cite{choudhury2015architecture}. Common challenges includes extracting vector images as they are embedded in the PDF document and extracting a large figure as a whole instead of multiple figures of its components. To this end, we have used a publicly existing tool called \textit{PDFFigures 2.0}\footnote{\url{https://github.com/allenai/pdffigures2}}~\cite{clark2016pdffigures} for extracting a list of figures from a scholarly paper. However, none of the existing open source tools maintain the table structure that is essential for us. Thus, we built a PDF table extraction tool using PDFMiner\footnote{\url{https://euske.github.io/pdfminer/}} and Poppler-utils\footnote{\url{https://poppler.freedesktop.org/}}. Poppler-utils provide high level information about the document such as the text dump, while using PDFMiner, certain low level document details such as vertical line spacing are obtained. The table structure, along with the table caption, is retrived by building the heuristics over the horizontal and vertical line spacing.


\subsection{Figure and Table Classification}
The aim is to classify and retrive only those figures and tables in a research paper that contains a DL design flow. Futher, a fine-grained classifier is required to classify the figure into one of the identified five broad categories and classify the table as a row-major or column-major flow. 

In case of figures, the classifier is trained to perform the prediction using the architecture shape and the flow. For example, figures having result graphs and showing sample images from dataset has different shape compared to an architecture flow diagram. All the figures are resized to $224\times224$ and $4,096$ features (\textit{fc2}) are extracted from a fully connected layer of a popular deep learning model \textit{VGG19}~\cite{simonyan2014very} pre-trained on ImageNet dataset. We have two classification levels: (i) Coarse classifier: a binary neural network (NNet) classifier trained on \textit{fc2} features to classify if the figure contains a DL model or not, and (ii) Fine-grained classifier: a five class neural network classifier trained on \textit{fc2} features to identify the type of DL design, only for those figures classified positive by the coarse classifier. Having a sequence of two classifiers provided better performance as compared to a single classifier with six classes (sixth class being no DL design flow).

In case of tables, a bag-of-words model is built using keywords from the caption text as well as the table text. A cosine distance based classifier is used to identify if there is a DL design flow in the given table as compared to tables containing results. Further based on the number of rows and columns in the table, as extracted in the previous section, the table is classified as a row-major or column-major flow. 

\subsection{Content Extraction from Figure}
\begin{figure}[!t]
	\centering
	\includegraphics[width=3.4in]{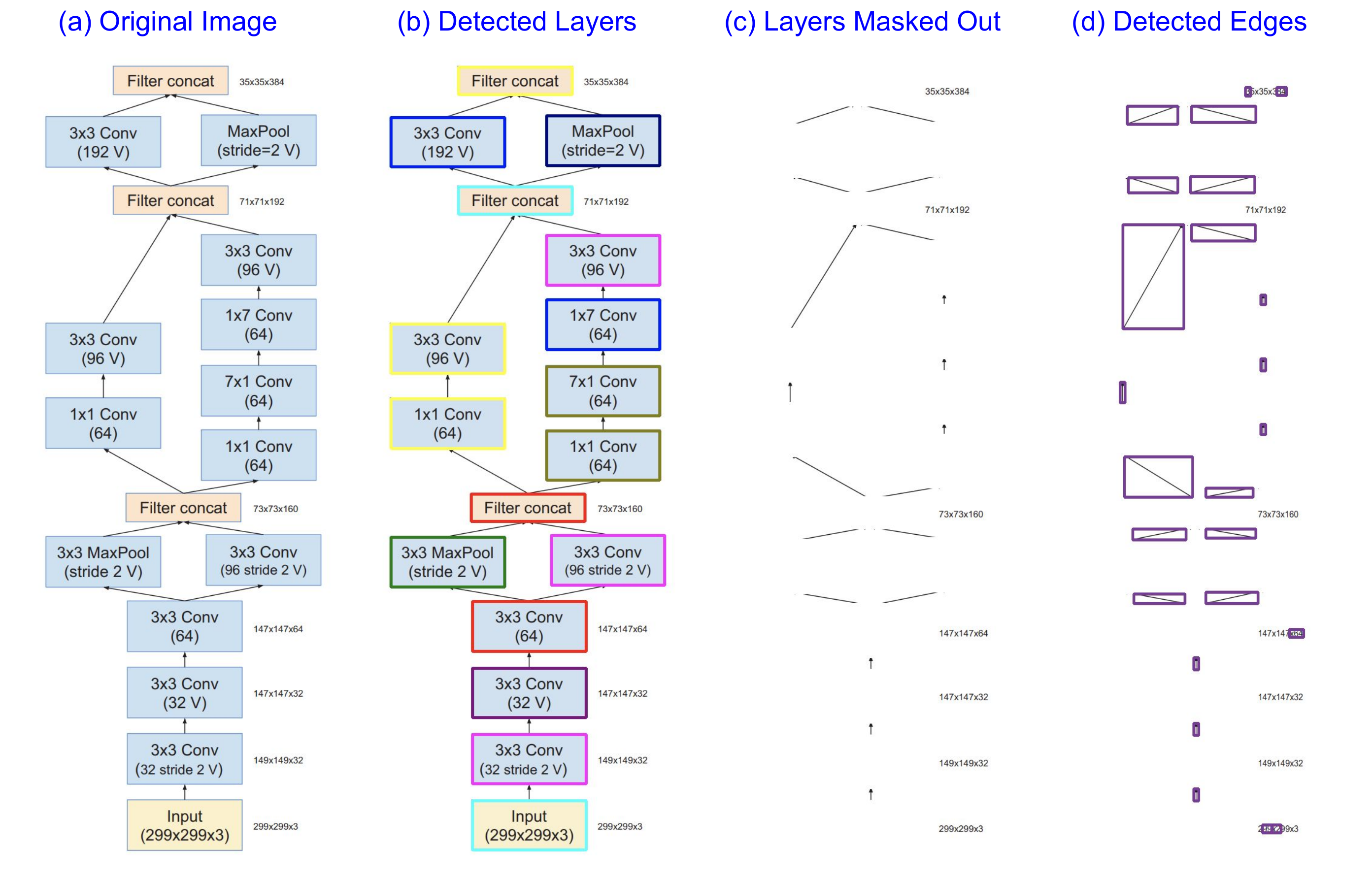}
	\caption{Illustration of the proposed flow detection approach from complex figures~\cite{szegedy2017inception} (AAAI 2017) involving (b) node/ layer detection, and (d) edge detection.}
	\label{fig:flow2}
\end{figure}

Content extraction from figures has two major steps: (i) Flow detection to identify the nodes and the edges, and (ii) OCR to extract the flow content. Identifying the flow is the challenges, as there is a huge variation in the type of DL design flow diagrams. In this section, we explain the details of the approach for a 2D Box type, as shown in Figure~\ref{fig:flow2}, while similar approach could be extended to other types, as well. Flow detection involves identifying the nodes first, followed by the edges connecting the nodes. As the image is usually of high resolution and quality, they are directly binarized using an adaptive Gaussian thresholding and a Canny edge detection approach is used to identify all the lines. An iterative region grown algorithm is adopted to identify closed countours in the figure, as they represent the nodes as shown in Figure~\ref{fig:flow2}(b). All the detected nodes are masked out from the figure and the contour detection algorithm is applied again to detection the edges,  as shown in Figure~\ref{fig:flow2}(d). The direction of the edge flow is obtained by analyzing the pixel distribution within each edge contour. The node and edge contours are then sorted based on the location and direction to obtain the flow of the entire DL model design. As shown in Figure~\ref{fig:flow2}, the proposed approach could also handle branchings and forking in a design flow diagram.  

Once the flow is extracted, the text in each node/ layer is obtained through OCR using Tesseract\footnote{\url{https://github.com/tesseract-ocr/}}.Based on our manual observation, we assume that the a layer description would be available within the detected node. A dictionary of possible DL layer names is created to perform spell correction of the extracted OCR text. 


\begin{figure}[!t]
	\centering
	\includegraphics[width=3.4in]{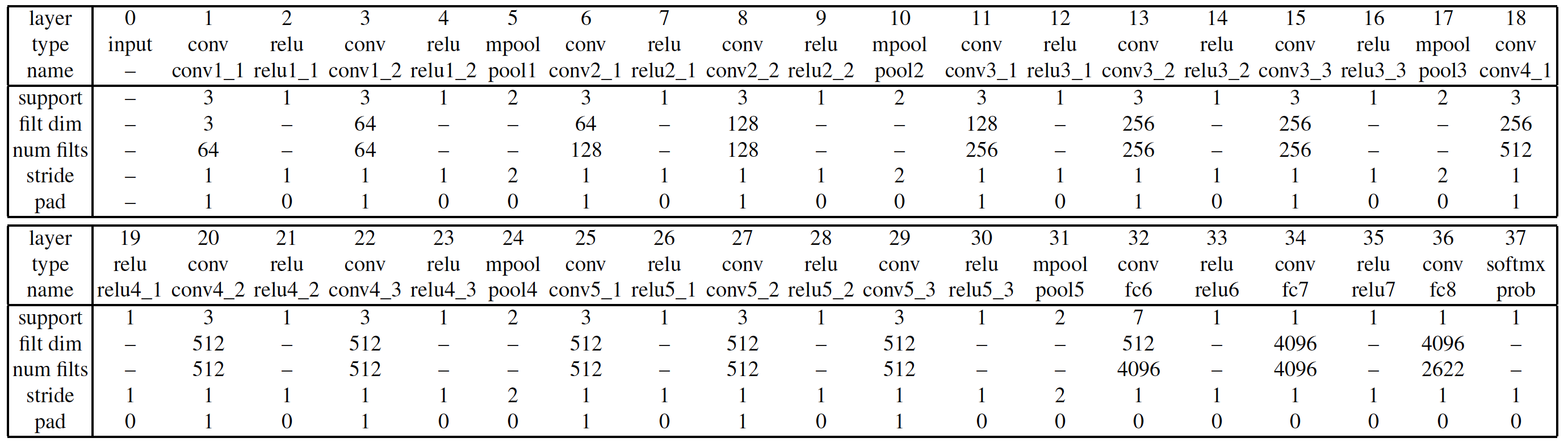}
	\caption{An example table showing the DL design flow as explained in tabular format in~\cite{parkhi2015deep}.}
	\label{fig:table}
\end{figure}

\subsection{Content Extraction from Table}
In a row major table, every row corresponds to a layer in the DL design flow, as shown in Figure~\ref{fig:table}. Similarly, in a column major table, every column corresponds to a layer along with other parameters of the layer. The layer name is extracted by matching it with a manually created dictionary. Further, the parameters are extracted by mapping the corresponding row or column header with a pre-defined list of parameter names corresponding to the layer. Thus, sequentially the entire DL design flow is extracted from a table.

\begin{figure}[!t]
	\centering
	\includegraphics[width=3.4in]{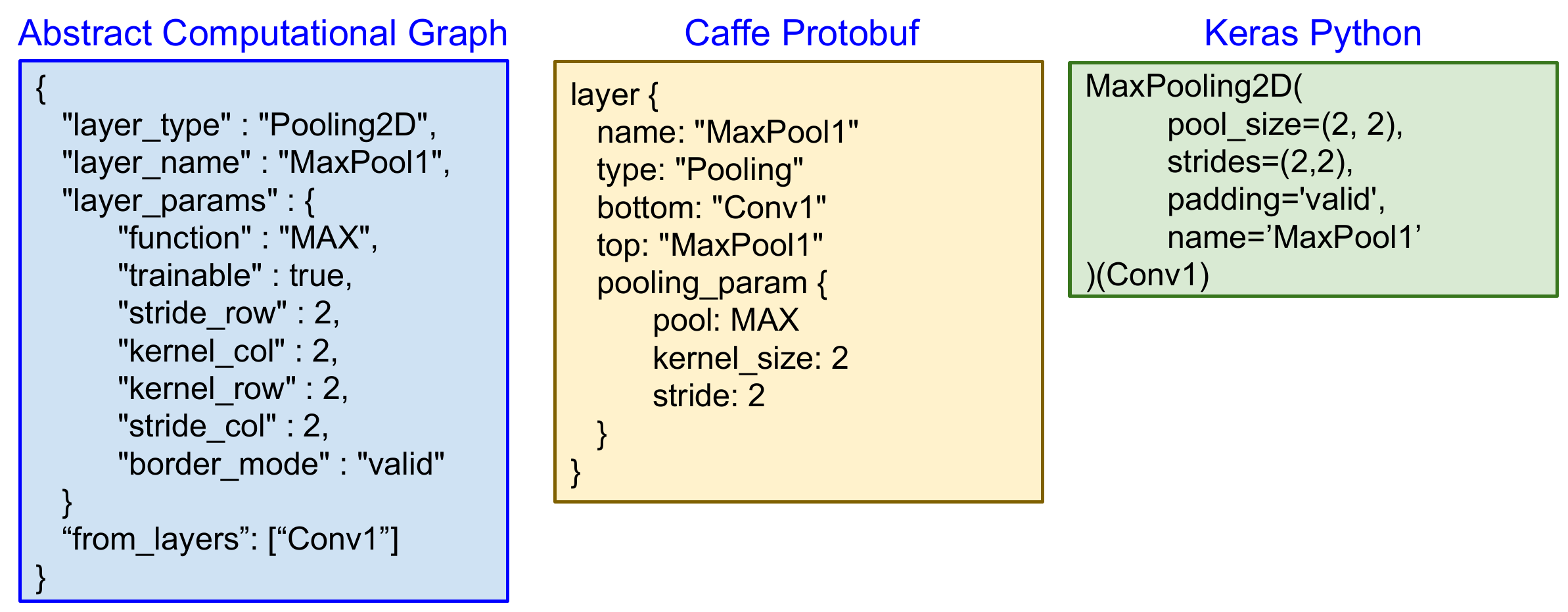}
	\caption{An illustration for a Pooling2D layer showing the rule base of the inference engine, converting the abstract JSON format into Caffe protobuf and Keras python code.}
	\label{fig:mapping}
\end{figure}

\subsection{Generating Source Code}
Overall, after detecting DL design flow, an abstract computational graph is represented in JSON format, as shown in Figure~\ref{fig:mapping}.
Two rule based converters are written to convert the abstract computational graph extracted in the previous step to either (i) Keras code (Python) or (ii) Caffe protobuf format (prototxt). An inference engine acts as the convertor to map the abstract computational graph to the grammar of the platform. The inference engine consists of a comprehensive list of templates and dictionaries built manually for both Keras and Caffe. Template based structures transfer each component from the abstract representation to a platform specific structure using the dictionary mappings. Further, another set of templates, consisting of a set of assertions, are designed for translating each layer's parameters. Further, the inference engine is highly flexible allowing easy extension and addition of new layer definitions. An example of the inference engine's dictionary mapping for a Pooling2D layer is shown in Figure~\ref{fig:mapping}.





Thus for a given research paper, by processing both the figure and table content, the DL model design flow is obtained which is converted to execution ready code in both Keras and Caffe. 

%% file: 3_Data_Simulation.tex
\begin{table*}[ht]
	\centering
	\begin{tabular}{|p{1.2cm}|p{0.8cm}|p{1.1cm}|p{1cm}|p{1.1cm}|p{1.1cm}|p{1.1cm}|p{0.9cm}|p{1cm}|p{.6cm}|p{.6cm}|p{0.9cm}|p{0.9cm}|}
		\hline
		\textbf{Current Layer} & \textbf{Dense} & \textbf{Conv2D} & \textbf{Flatten} & \textbf{Dropout} & \textbf{MaxPool} & \textbf{AvgPool} & \textbf{Concat} & \textbf{Embed}  & \textbf{RNN} & \textbf{RNN (seq)} & \textbf{LSTM} & \textbf{LSTM (seq)}\\ \hline
		
		\textbf{Input}  & \ding{52} & \ding{52} &  &  &  &  &  & \ding{52} &  &  &  & \\ \hline
		
		\textbf{Dense} & \ding{52} &  &  & \ding{52} &  &  & \ding{52} & \ding{52} &  &  &  & \\ \hline
		\textbf{Conv2D} &  & \ding{52} & \ding{52} & \ding{52} & \ding{52} & \ding{52} & \ding{52} &  &  &  &  & \\ \hline
		\textbf{Flatten} & \ding{52} &  &  & \ding{52} &  &  & \ding{52} & \ding{52} &  &  &  & \\ \hline
		\textbf{Dropout} &  \multicolumn{12}{|c|}{Same as previous layer} \\ \hline
		\textbf{MaxPool} &  & \ding{52} & \ding{52} & \ding{52} & \ding{52} & \ding{52} & \ding{52} &  &  &  &  & \\ \hline
		\textbf{AvgPool} &  & \ding{52} & \ding{52} & \ding{52} & \ding{52} & \ding{52} & \ding{52} &  &  &  &  & \\ \hline
		\textbf{Concat} & \multicolumn{12}{|c|}{If input is one dimensional, same as Dense layer; else same as previous layer} \\ \hline
		\textbf{RNN} & \ding{52} &  &  & \ding{52} &  &  &  \ding{52}& \ding{52} &  &  &  & \\ \hline
		\textbf{RNN (seq)} &  &  & \ding{52} & \ding{52} &  &  & \ding{52} &  & \ding{52} & \ding{52} & \ding{52} &\ding{52} \\ \hline
		\textbf{LSTM} && \ding{52}  &  & \ding{52} &  &  &  \ding{52}& \ding{52} &  &  &  & \\ \hline
		\textbf{LSTM (seq)}  &  &  & \ding{52} & \ding{52} &  &  & \ding{52} &  & \ding{52} & \ding{52} & \ding{52} &\ding{52} \\ \hline
		
	\end{tabular}
	\caption{\label{tab:table1} The proposed grammar for creating valid deep learning design models defining the list of possible next layers for a given current layer.}
\end{table*}

\begin{table}[t]
	\centering
	\begin{tabular}{|c|c|}
		\hline
		\textbf{Layer} & \textbf{Hyper-parameters} \\ \hline
		
		\multirow{1}{*}{Dense} & \#nodes - \{[5:5:500]\}  \\ \hline 
		\multirow{1}{*}{Dropout} & probability - \{[0:0.1:1]\}  \\ \hline 
		\multirow{2}{*}{Conv2D} & \#filters - \{[16:16:256]\}  \\ \cline{2-2} 
		& filter size - \{[1:2:11]\}  \\ \hline
		\multirow{2}{*}{MaxPool} & stride - \{[2:1:5]\}  \\ \cline{2-2} 
		& filter size - \{[1:2:11]\}  \\ \hline
		\multirow{2}{*}{AvgPool} & stride - \{[2:1:5]\}  \\ \cline{2-2} 
		& filter size - \{[1:2:11]\}  \\ \hline 
		\multirow{2}{*}{Embed} & embed size - \{64, 100, 128, 200\}  \\ \cline{2-2} 
		& vocab - \{[10000, 20000, 50000, 75000]\}  \\ \hline
		\multirow{1}{*}{SimpleRNN} & \#units - \{[3:1:25]\}  \\ \hline 
		\multirow{1}{*}{LSTM} & \#nodes - \{[3:1:25]\}  \\ \hline 
		\multirow{4}{*}{InputData} & MNIST - \{28, 28, 1\}  \\ \cline{2-2} 
		& CIFAR-10 - \{32, 32, 3\}  \\ \cline{2-2} 
		& ImageNet - \{224, 224, 3\}  \\ \cline{2-2} 
		& IMDB Text  \\ \hline
		
	\end{tabular}
	\caption{\label{tab:table2} The set of hyper-parameter options considered for each layer in our simulated dataset generation. The parameter value [a:b:c] means a list of values from $a$ to $c$ in steps of $b$.}
\end{table}

\section{Evaluation on Simulated Dataset}

The aim of this process is to simulate and generate ground truth deep learning designs and their corresponding flow visualizations figures. Thus, the proposed pipeline of DL model design could be quantitatively evaluated. To this end, we observed that both Keras and Caffe have an in-built visualization routine for a DL design model. Further both Keras and Caffe have their internal DL model validator and a visualization can be exported only when the simulated design is deemed valid by the validator.

\subsection{Grammar for DL Design Generation} 
To be able to generate meaningful and valid DL design models, we manually defined a grammar for the model flow as well as for the hyper-parameters. We considered $10$ unique layers for our dataset simulation - \{\textit{Conv2D, MaxPool2D, AvgPool2D}\} for building convolution neural network like architectures, \{\textit{Embed, SimpleRNN, LSTM}\} for building recurrent neural network like architectures, \{\textit{Dense, Flatten, Dropout, Concat}\} as the core layers. The use of \textit{Concat} enables our designed models to be non-sequential as well as with a combination of recurrent and convolution architectures. This allows us to create  random, complex, and highly varying DL models. Also, \textit{RNN} and \textit{LSTM} layers have an additional binary parameter of \textit{return seq}, which when set true returns the output of every hidden cell, otherwise, returns the output of only the last hidden cell in the layer. 
Table~\ref{tab:table1} explains the proposed grammar for generating DL design models. The grammar defines the set of all possible next layers for a given current layer. This is determined by the shape of the tensor flowing through each of the layer's operation. For example, a \textit{Dense} layer strictly expect the input to be a vector of shape $n\times 1$. Thus, the \textit{Dense} cannot appear after a \textit{Conv2D} layer without the presence of a \textit{Flatten} layer. The proposed grammar further includes the set of possible values for each hyper-parameter of a layer, as explained in Table~\ref{tab:table2}. While hyper-parameter values beyond the defined bounds are possible, the table values indicate the assumed set of values in the model simulation process.


\subsection{Simulated Dataset}
A model simulation starts with an \textit{Input} layer, where there are four possible options - \textit{MNIST, CIFAR, ImageNet, IMDBText}. From the set of all possible next layers for the given \textit{Input} layer, a completely random layer is decided. For the given next layer, a random value is picked for every possible hyper-parameter. For example, for \textit{MNIST} being the input layer, \textit{Conv2D} could be picked as the random next layer. Then, for \textit{Conv2D} the hypar-parameters are determined randomly as $32$ filters, $5\times5$ filter size with a stride of $2$. The model design always ends with a \textit{Dense} layer with number of nodes equal to the number of classes of the corresponding \textit{Input} layer.

The number of layers in between the \textit{Input} layer and the final \textit{Dense} layer denotes the depth of the DL model. For our simulation, we generated $3,000$ DL models for each depth varied between $5$ and $40$, creating a total dataset of $108,000$ models. Each model contains the Keras JSON representation, Keras image visualization, Caffe protobuf files, and Caffe image visualization, resulting in a total of $216,000$ DL model design visualizations. These models are valid by construct since they follow a well-defined grammar. However, these models need not be the best from an execution perspective, or with respect to their training performance. 

\begin{table}[t]
	\centering
	\begin{tabular}{|l|c|c|c|}
		\hline
		\textbf{Observation} & \textbf{Train} & \textbf{Validation}  & \textbf{Test} \\ \hline
		
		\#Points & $195,296$ & $48,824$ & $48,824$ \\ \hline			
		Naive Bayes & $98.29\%$ & $98.30\%$ & $98.39\%$ \\ \hline
		Decision Tree & $100\%$ & $99.57\%$ & $99.55\%$ \\ \hline
		Logistic Regression & $100\%$ & $99.98\%$ & $99.99\%$ \\ \hline
		RDF & $100\%$ & $99.72\%$ & $99.68\%$ \\ \hline
		SVM (RBF Kernel) & $100\%$ & $99.89\%$ & $99.83\%$ \\ \hline
		Neural Network & $100\%$ & $99.93\%$ & $99.94\%$ \\ \hline
	\end{tabular}
	\caption{\label{tab:table3} The performance of various binary classifiers to distinguish KerasCaffeVisulizations vs. other often occuring images in research papers.}
\end{table}

\subsection{Figure Type Classification Performance}
In this experiment, a binary NNet classifier with two hidden layers of size [$1024$, $256$] is trained on \textit{fc2} features of \textit{VGG19} model to differentiate $216,000$ simulated DL visualizations from a set of $28,120$ other kind of diagrams often available in research papers (scraped from PDF). The whole dataset is split as $60\%$ for training, $20\%$ validation, and $20\%$ for testing, making it a total of $195,296$ images for training and $48,824$ images for testing. The performance of the NNet classifier is compared with six different classifiers as shown in Table~\ref{tab:table3}. As it can be observed most of the classifier provide a classification accuracy of $100\%$, showing that from a set of figures obtained from a research paper, it would be possible to distinguish only the deep learning design flow diagrams. All the classifiers use the default parameters as provided the \textit{scikit-learn} package.

\subsection{Computational Graph Extraction Performance}
In this experiment, the performance of flow and content extraction from the $216,000$ Keras and Caffe visualizations is evaluated against the ground truth. By performing OCR, on the extracted flow, the unique layer names are obtained and two detection accuracies are reported,
\begin{enumerate}
	\item blob (or layer) detection accuracy: evaluates the performance of blob detection and layers identified using OCR and is computed as the ratio of correct blobs detected per model (in percent)
	\item edge detection accuracy: evaluates the performance of the detected flow and is computed as the ratio of correct arrows detected per model (in percent)
\end{enumerate}

\noindent Figure~\ref{fig:boxplot} is the box plot showing the performance for the the proposed figure extraction pipeline in both Keras and Caffe. As it can be observed, the proposed pipeline provides $100\%$ accuracy in layer extraction and more than $93\%$ accuracy in extracting the edges. As the edges can be curved and can be of any length, even connecting the first with the last layer, the variations caused a reduction in performance. 
Further details on extracting flow information Keras and Caffe visuazliation and also for additional results, kindly refer to the supplementary material.

%% file: 4_Experimental_Study.tex
\begin{figure}[t]
	\centering
	\includegraphics[width=3.4in]{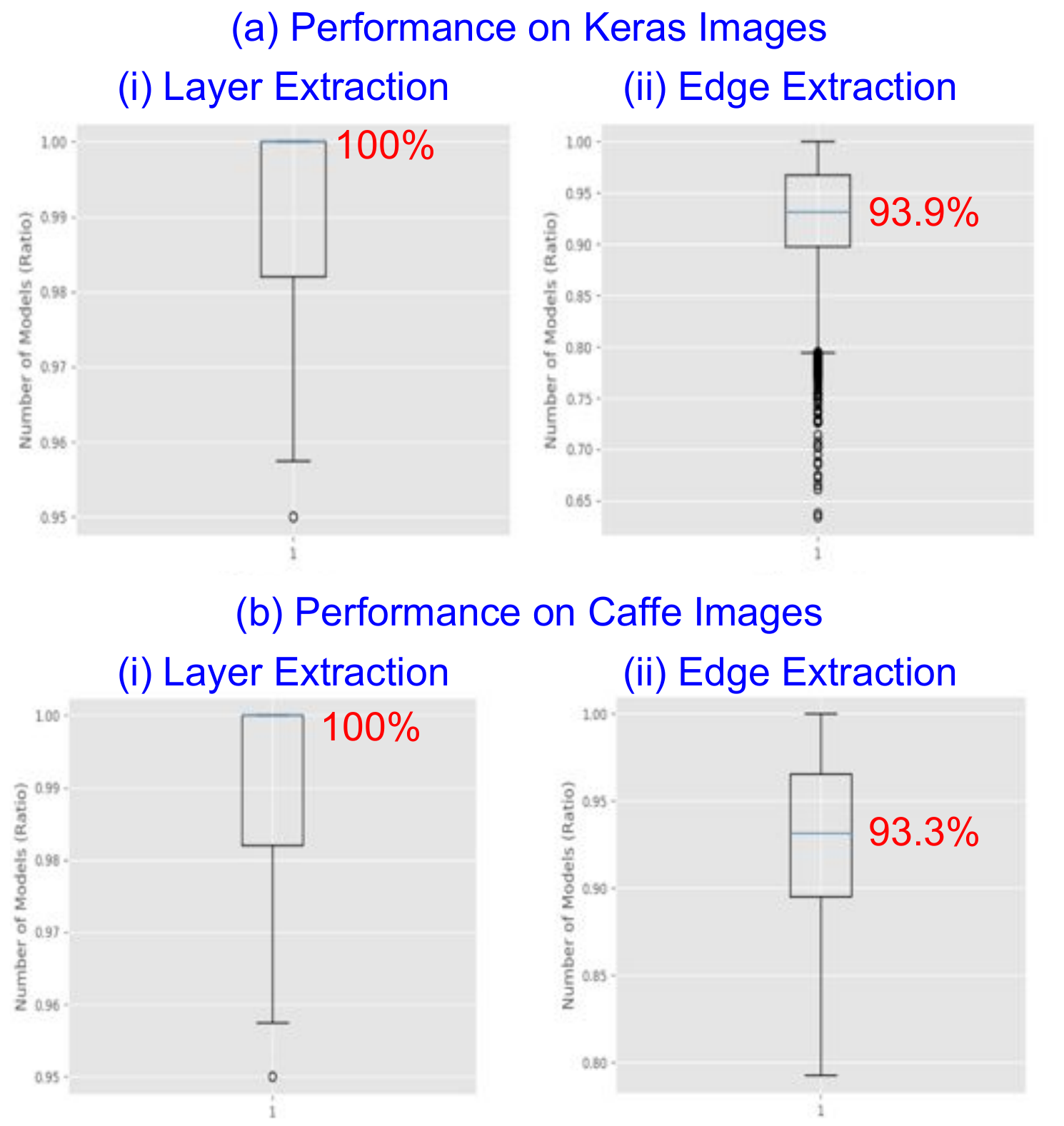}
	\caption{Box plots showing the performance accuracy of flow detection in Keras and Caffe visualizations.}
	\label{fig:boxplot}
\end{figure}

\section{Results on Deep Learning Scholarly Papers}

The first $5,000$ papers were downloaded from \textit{arXiv.org} using \textit{``deep learning"} as the input query. $30,987$ figures were extracted from these $5,000$ downloaded papers, out of which $28,120$ figures did not contain a DL design flow while the remaining $2,867$ contained. These represent the usual figures that are found in a deep learning research paper that does not contain a design flow. 

\subsection{Figure Type Classification Accuracy}
To evaluate the coarse level binary classification, a 2 hidden layer NNet was trained on the \textit{fc2} features obtained from the $30,987$ images extracted from research papers. The whole dataset is split as $60\%$ for training, $20\%$ validation, and $20\%$ for testing and the results are computed for seven different classifiers as shown in Table~\ref{tab:table4}.

Further, to evaluate the fine level, five-class, figure type classification, the $2,871$ DL design flow diagrams were manually labelled. The distribution of figures were as follows: (i) Neurons plot: $587$ figures, (ii) 2D box: $1,204$, (iii) Stacked2D box: $408$, (iv) 3D box: $562$, and (v) Pipeline plot: $110$. A $60-20-20$ train, validation, and test split is performed to train the NNet classifier in comparison with the six other classifier to perform this five class classification. The results are table in Table~\ref{tab:table5}. The results show that even on highly varying DL flow design images, identifying the type of DL flow is more than $70\%$ accurate. For more details on the experimental analysis, please refer to the supplementary material.

\begin{table}[t]
	\centering
	\begin{tabular}{|l|c|c|c|}
		\hline
		\textbf{Observation} & \textbf{Train} & \textbf{Validation}  & \textbf{Test} \\ \hline
		
		\#Points & $18,592$ & $6,197$ & $6,198$ \\ \hline			
		Naive Bayes & $77.29\%$ & $64.39\%$ & $62.56\%$ \\ \hline
		Decision Tree & $99.96\%$ & $76.67\%$ & $74.35\%$ \\ \hline
		Logistic Regression & $99.96\%$ & $86.17\%$ & $85.27\%$ \\ \hline
		RDF & $99.96\%$ & $83.72\%$ & $82.94\%$ \\ \hline
		SVM (RBF Kernel) & $99.96\%$ & $86.89\%$ & $85.25\%$ \\ \hline
		Neural Network & $99.96\%$ & $87.93\%$ & $86.25\%$ \\ \hline
	\end{tabular}
	\caption{\label{tab:table4} The performance of coarse level binary classifier to distinguish DL design flow figures from other figures that usually appear in a research paper.}
\end{table}

\begin{table}[t]
	\centering
	\begin{tabular}{|l|c|c|c|}
		\hline
		\textbf{Observation} & \textbf{Train} & \textbf{Validation}  & \textbf{Test} \\ \hline
		
		\#Points & $1,720$ & $573$ & $574$ \\ \hline			
		Naive Bayes & $40.42\%$ & $54.30\%$ & $52.84\%$ \\ \hline
		Decision Tree & $99.65\%$ & $50.57\%$ & $49.13\%$ \\ \hline
		Logistic Regression & $99.65\%$ & $69.98\%$ & $68.47\%$ \\ \hline
		RDF & $99.65\%$ & $68.72\%$ & $66.02\%$ \\ \hline
		SVM (RBF Kernel) & $99.65\%$ & $72.94\%$ & $69.68\%$ \\ \hline
		Neural Network & $100\%$ & $74.93\%$ & $71.60\%$ \\ \hline
	\end{tabular}
	\caption{\label{tab:table5} The performance of fine level, five class classifier to identify the type of DL design flow figure obtained from the research paper.}
\end{table} 

\begin{figure}[t]
	\centering
	\includegraphics[width=3.4in]{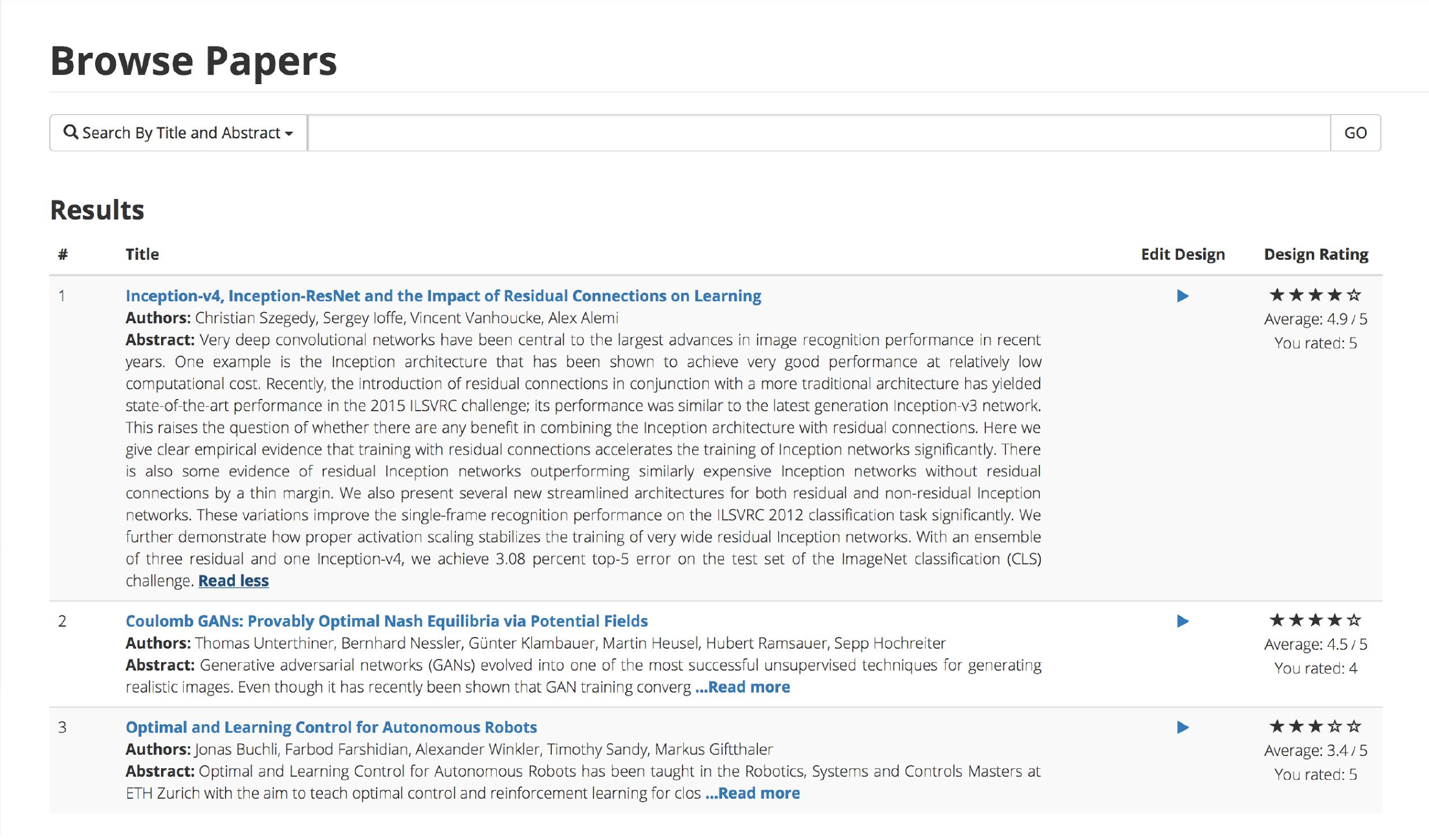}
	\caption{An \textit{arXiv}-like website where DL papers along with their extracted design, and generated source code in Caffe and Keras is made available.}
	\label{fig:arxivpaper}
\end{figure}

\begin{figure}[t]
	\centering
	\includegraphics[width=3.4in]{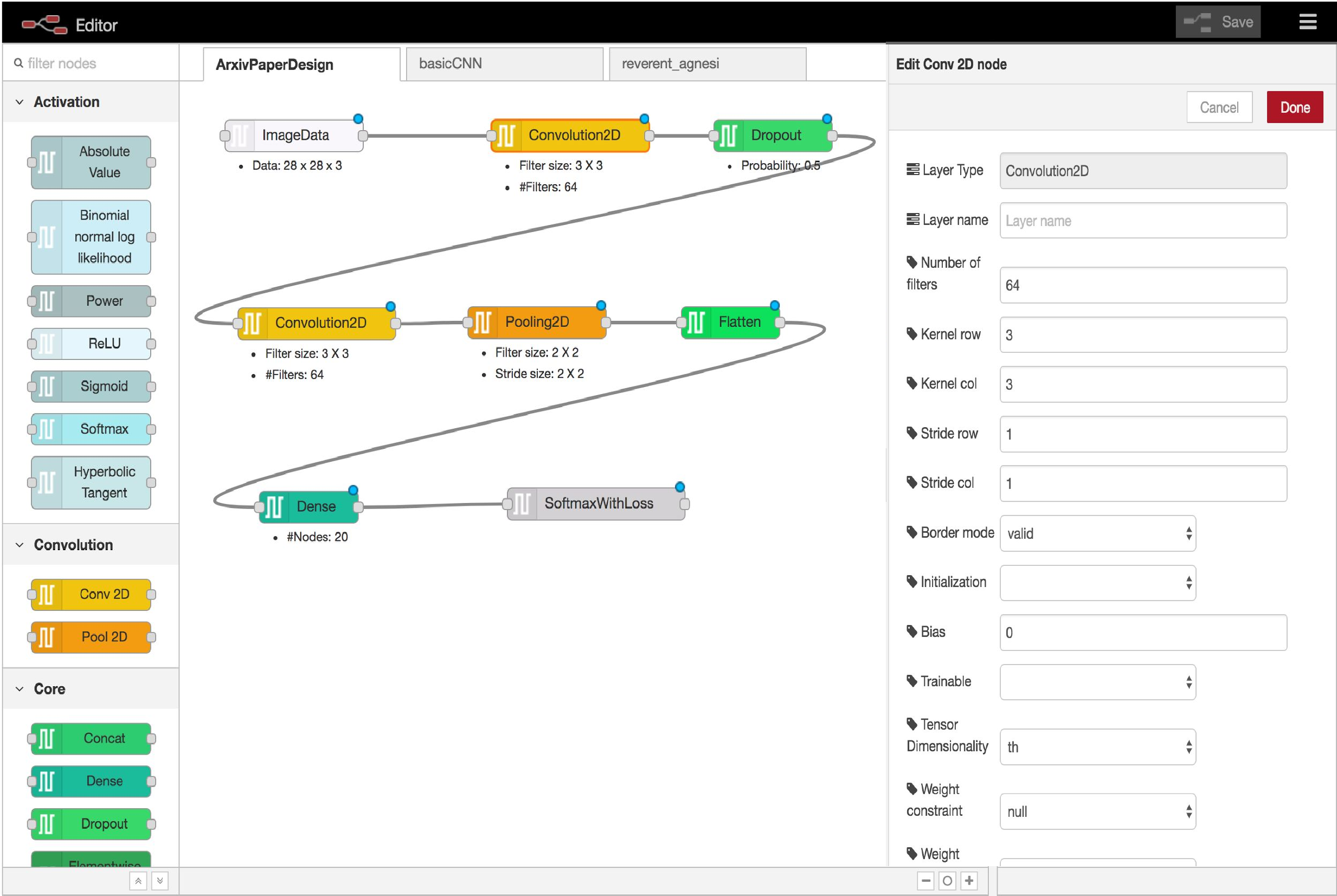}
	\caption{An intuitive drag-and-drop UI based framework to edit the extracted designs and make them publicly available.}
	\label{fig:arxivdesign}
\end{figure}

\subsection{Crowdsourced Improvement of Extracted Designs}
Using the proposed DLPaper2Code framework, we extracted the DL design models for all the $5,000$ downlaoded papers. However, quantitatively evaluating the extracted design flow would be challenging due to the lack of a ground truth. Hence, we created an \textit{arXiv}-like website, as shown in Figure~\ref{fig:arxivpaper}, where the papers, the corresponding design, and the generated source code is available. The DL community could rate the extracted designs which acts as a feedback or performance measure of our automated approach.

Further, an intuitive drag-and-drop based UI framekwork is generated for the community to edit the generated DL flow design, as shown in Figure~\ref{fig:arxivdesign}. Ideally the respective papers' author or the DL community could edit the generated designs, wherever an error was found. The edited design could be further made publicly available for other researcher to reproduce their design. Further our system could generate the source code of the edited design in both Keras and Caffe, in real-time. Thus, we have a two-fold advantange through this UI system: (i) the public system could act as a one-stop repository for any DL paper and it's corresponding design flow and source code, (ii) the feedback the community provides would enable us to continuously learn and improve the system\footnote{To adhere to the double-blind submission policy of AAAI, the system URL is not provided in this version of the paper}.

%% file: 7_Conclusion.tex
\section{Conclusion and Discussion}
Thus, researchers and developers need not struggle any further in reproducing research papers in deep learning. Using this research, the DL design explained in a research paper could be automatedly extracted. Using an intuitive drag-and-drop based UI editor, that we designed as a part of this research, the extracted design could be manually edited and perfected. Further for an extracted DL design, the source code could be generated in Keras (Python) and Caffe (prototxt), in real-time. The proposed DLpaper2Code framework extracts both figure and table information from a research paper and converts it into source code. Currently, an \textit{arXiv}-like website is created that contains the DL design and the source code for $5,000$ research papers. To evaluate our approach, we simulated a dataset of $108,000$ unique deep learning designs validated by a proposed grammar and their corresponding Keras and Caffe visualizations. On a total dataset of $216,000$ deep learning model visualization diagrams and $28,120$ diagrams that appeared in deep learning research papers and did not contains a model visualization, the proposed binary classification using NNet classifier obtained $99.9\%$ accuracy. The performance of extracting the generic computational graph figures using the proposed pipeline is more than $93\%$.
While this research could have a high impact in the reproducibility of DL research, we have planned for plenty of possible extensions for the proposed pipeline:
\begin{enumerate}
	\item The proposed pipeline detects only the layers (blobs) and the edges from the diagram. It could be extended to detect and extract the hyper-parameter values of each layer, to make the computational graph more content rich.
	\item Currently, we have two independent pipelines for generating abstract computational graphs from tables and figures. Combining the information obtained from the multi-modal sources could enhance the accuracy of the extracted DL design flow.
	\item The entire DLPaper2Code framework could be extended to support additional libraries, apart from Keras and Caffe, such as Torch, Tensorflow etc.
	\item The broader aim would be to propose a definition of representating DL model design in research papers, achieving uniformity and better readibility. Further, authors of future papers could also release their design in the created website for easy accessibility to the community.
\end{enumerate}

%% file: main.bbl
\begin{thebibliography}{}

\bibitem[\protect\citeauthoryear{Abadi \bgroup et al\mbox.\egroup
  }{2016}]{tensorflow2015}
Abadi, M.; Agarwal, A.; Barham, P.; Brevdo, E.; Chen, Z.; Citro, C.; Corrado,
  G.~S.; Davis, A.; Dean, J.; Devin, M.; et~al.
\newblock 2016.
\newblock Tensorflow: Large-scale machine learning on heterogeneous distributed
  systems.
\newblock {\em arXiv preprint arXiv:1603.04467}.

\bibitem[\protect\citeauthoryear{Bastien \bgroup et al\mbox.\egroup
  }{2012}]{theano2012}
Bastien, F.; Lamblin, P.; Pascanu, R.; Bergstra, J.; and Goodfellow, I. J.
  e.~a.
\newblock 2012.
\newblock Theano: new features and speed improvements.
\newblock Deep Learning and Unsupervised Feature Learning NIPS 2012 Workshop.

\bibitem[\protect\citeauthoryear{Chen}{2015}]{chen2015mxnet}
Chen, T. e.~a.
\newblock 2015.
\newblock Mxnet: A flexible and efficient machine learning library for
  heterogeneous distributed systems.
\newblock {\em arXiv preprint arXiv:1512.01274}.

\bibitem[\protect\citeauthoryear{Chintala}{2016}]{pytorch}
Chintala, S.
\newblock 2016.
\newblock Pytorch.
\newblock \url{https://github.com/pytorch/pytorch}.

\bibitem[\protect\citeauthoryear{Chollet and others}{2015}]{chollet2015keras}
Chollet, F., et~al.
\newblock 2015.
\newblock Keras.
\newblock \url{https://github.com/fchollet/keras}.

\bibitem[\protect\citeauthoryear{Choudhury and
  Giles}{2015}]{choudhury2015architecture}
Choudhury, S.~R., and Giles, C.~L.
\newblock 2015.
\newblock An architecture for information extraction from figures in digital
  libraries.
\newblock In {\em WWW (Companion Volume)},  667--672.

\bibitem[\protect\citeauthoryear{Clark and Divvala}{2016}]{clark2016pdffigures}
Clark, C., and Divvala, S.
\newblock 2016.
\newblock Pdffigures 2.0: Mining figures from research papers.
\newblock In {\em Digital Libraries (JCDL), 2016 IEEE/ACM Joint Conference on},
   143--152.

\bibitem[\protect\citeauthoryear{Dieleman}{2015}]{lasagne}
Dieleman, S.
\newblock 2015.
\newblock Lasagne: First release.

\bibitem[\protect\citeauthoryear{Donahue \bgroup et al\mbox.\egroup
  }{2015}]{donahue2015long}
Donahue, J.; Anne~Hendricks, L.; Guadarrama, S.; Rohrbach, M.; Venugopalan, S.;
  Saenko, K.; and Darrell, T.
\newblock 2015.
\newblock Long-term recurrent convolutional networks for visual recognition and
  description.
\newblock In {\em Computer vision and pattern recognition},  2625--2634.

\bibitem[\protect\citeauthoryear{et al}{2011}]{torch}
et~al, R.~C.
\newblock 2011.
\newblock Torch7: A matlab-like environment for machine learning.
\newblock In {\em BigLearn, NIPS Workshop}.

\bibitem[\protect\citeauthoryear{Gibson}{2015}]{DL4J}
Gibson, A.
\newblock 2015.
\newblock Dl4j.
\newblock \url{https://github.com/deeplearning4j/deeplearning4j}.

\bibitem[\protect\citeauthoryear{Jia \bgroup et al\mbox.\egroup
  }{2014}]{jia2014caffe}
Jia, Y.; Shelhamer, E.; Donahue, J.; Karayev, S.; Long, J.; Girshick, R.;
  Guadarrama, S.; and Darrell, T.
\newblock 2014.
\newblock Caffe: Convolutional architecture for fast feature embedding.
\newblock In {\em ACM international conference on Multimedia},  675--678.

\bibitem[\protect\citeauthoryear{Karpathy and Fei-Fei}{2015}]{karpathy2015deep}
Karpathy, A., and Fei-Fei, L.
\newblock 2015.
\newblock Deep visual-semantic alignments for generating image descriptions.
\newblock In {\em Computer Vision and Pattern Recognition},  3128--3137.

\bibitem[\protect\citeauthoryear{Parkhi \bgroup et al\mbox.\egroup
  }{2015}]{parkhi2015deep}
Parkhi, O.~M.; Vedaldi, A.; Zisserman, A.; et~al.
\newblock 2015.
\newblock Deep face recognition.
\newblock In {\em BMVC}, volume~1, ~6.

\bibitem[\protect\citeauthoryear{Sankaran \bgroup et al\mbox.\egroup
  }{2011}]{sankaran}
Sankaran, A.; Aralikatte, R.; Mani, S.; Khare, S.; Panwar, N.; and Gantayat, N.
\newblock 2011.
\newblock {DARVIZ}: Deep abstract representation, visualization, and
  verification of deep learning models: Nier track.
\newblock In {\em International Conference on Software Engineering},  804--807.

\bibitem[\protect\citeauthoryear{Seide and Agarwal}{2016}]{CNTK}
Seide, F., and Agarwal, A.
\newblock 2016.
\newblock Cntk: Microsoft's open-source deep-learning toolkit.
\newblock In {\em Proceedings of the 22Nd ACM SIGKDD International Conference
  on Knowledge Discovery and Data Mining}, KDD '16,  2135--2135.
\newblock New York, NY, USA: ACM.

\bibitem[\protect\citeauthoryear{Simonyan and
  Zisserman}{2014}]{simonyan2014very}
Simonyan, K., and Zisserman, A.
\newblock 2014.
\newblock Very deep convolutional networks for large-scale image recognition.
\newblock {\em arXiv preprint arXiv:1409.1556}.

\bibitem[\protect\citeauthoryear{Springenberg \bgroup et al\mbox.\egroup
  }{2014}]{springenberg2014striving}
Springenberg, J.~T.; Dosovitskiy, A.; Brox, T.; and Riedmiller, M.
\newblock 2014.
\newblock Striving for simplicity: The all convolutional net.
\newblock {\em arXiv preprint arXiv:1412.6806}.

\bibitem[\protect\citeauthoryear{Szegedy \bgroup et al\mbox.\egroup
  }{2017}]{szegedy2017inception}
Szegedy, C.; Ioffe, S.; Vanhoucke, V.; and Alemi, A.~A.
\newblock 2017.
\newblock Inception-v4, inception-resnet and the impact of residual connections
  on learning.
\newblock In {\em AAAI},  4278--4284.

\bibitem[\protect\citeauthoryear{Vinyals \bgroup et al\mbox.\egroup
  }{2015}]{vinyals2015show}
Vinyals, O.; Toshev, A.; Bengio, S.; and Erhan, D.
\newblock 2015.
\newblock Show and tell: A neural image caption generator.
\newblock In {\em Computer Vision and Pattern Recognition},  3156--3164.

\end{thebibliography}
